\documentclass[10pt,journal]{IEEEtran}
\usepackage{amsmath,amssymb,graphicx,booktabs,cite,multirow,array,microtype}
\usepackage{dblfloatfix}
\usepackage{float}
\usepackage{url}
\usepackage{balance}
\begin{document}
\title{Task-Conditional Faithfulness Auditing of Multimodal LLMs for Grid Diagnosis}
\author{Tianqiao~Zhao,~\IEEEmembership{Senior Member,~IEEE}, Meng~Yue,~\IEEEmembership{Member,~IEEE}, and Jianhui~Wang,~\IEEEmembership{Fellow,~IEEE}%
\thanks{T. Zhao is with The University of Texas at Arlington, Arlington, TX, USA. M. Yue is with Brookhaven National Laboratory, Upton, NY, USA. J. Wang is with Southern Methodist University, Dallas, TX, USA.}
\vspace{-2em}}
\maketitle

\begin{abstract}
Multimodal large language models (LLMs) can combine topology, measurements, and incident text for grid diagnosis, yet answer accuracy does not establish that task-appropriate evidence was used. This letter proposes a general framework in order to conduct task-conditional faithfulness audit. It compares self-reported reliance, intervention-derived behavioral reliance, and preregistered engineering importance. The framework first registers task-specific evidence requirements and compares them with self-reported reliance and behavioral changes under controlled modality ablations. To resolve detected discrepancies, we design an evidence-gated correction and re-audit mechanism that regenerates failed responses under evidence constraints and independently re-ablates them to verify improved grounding without performance loss. Case studies evaluate three differently scaled LLMs on IEEE 39- and 118-bus scenarios. These results validate the framework ability to detect, diagnose, and correct task-conditional faithfulness failures.
\end{abstract}

\begin{IEEEkeywords}
Explainable AI, large language models, model faithfulness, multimodal fusion, power systems.
\end{IEEEkeywords}
\vspace{-0.6cm}
\section{Introduction}
LLM-based grid decision support is emerging across power dispatch,
numeric anomaly detection, and multimodal energy-management-system
security \cite{gaia_dispatch,llm_anomaly,grid_security}. Related
power-system fault-diagnosis studies also demonstrate the value of
combining heterogeneous measurements rather than relying on a single
information source \cite{transformer_fusion}. Multimodal availability,
however, does not guarantee task-appropriate evidence use. A correct
event label may be copied from an incident narrative even when
branch-status or dynamic evidence is ignored. This matters because
topology identification, security screening, and impact estimation
require different evidence pathways. More broadly, plausible model
explanations may not reflect the factors that drive a prediction, and
multimodal models can over-rely on one modality
\cite{turpin,chen_bias}.

This letter asks whether a multimodal grid LLM uses the evidence required by each task and how discrepancies should be corrected. It contributes: 1) a preregistered task contract that separates engineering evidence requirements from model explanations; 2) a validity-aware SR--Abl--Ref audit combining responsiveness and signed modality effects; 3) a diagnosis and corrective re-audit gate that requires measurable behavioral improvement without material performance loss; and 4) a generic implementation pipeline that transfers across grid systems and LLMs through lightweight system and model adapters. A three-model IEEE 39/118 proof of concept applies one frozen audit contract and reports output validity explicitly so that malformed intervention sets cannot inflate faithfulness.
\vspace{-0.4cm}
\section{Methodology}
\subsection{Problem Formulation}
Let $x_n=(m_{n,1},\ldots,m_{n,K},z_n)$ denote scenario $n$, with modality $m_{n,i}$ and operating metadata $z_n$; here $K=3$ for topology ($T$), measurements ($S$), and incident text ($X$). For task $j$, model $F_\theta$ returns a structured response $R_{j,n}$ containing answer $\hat y_{j,n}$ and reported reliance $s_{j,n}\in\Delta^{K-1}$. A task score $p_j(\hat y,y)\in[0,1]$ is oriented so that larger is better. The audit object is
\begin{equation}
\mathcal A_{j,n}=\{p_{j,n},s_{j,n},b_{j,n},g_j(x_n),V_{j,n},\mathbf I_{j,n}\},
\label{eq:audit_object}
\end{equation}
where $b$ is observable reliance, $g$ is registered engineering importance, $V$ indicates a valid output, and $\mathbf I$ records intervention validity. Thus, predictive correctness, behavioral grounding, and explanation fidelity are evaluated separately. The framework uses only observable input--output interventions; it does not require access to hidden chain-of-thought or treat self-reports as causal explanations. Aggregate alignment is reported jointly with $P_j$, validity, and responsiveness so that a model cannot appear faithful by failing to answer difficult cases.

\subsection{Task-Conditioned Reference Registration}
Before any test query, register
\begin{equation}
\mathcal T_j=(q_j,o_j,h_j,\mathcal Y_j,p_j,d_j,\mathcal E_j,\mathcal L_j,g_j),
\label{eq:task}
\end{equation}
where $q_j$ is the query; $o_j,h_j$ parse the answer and reliance; $d_j$ measures response change; $\mathcal E_j$ contains admissible evidence; and $\mathcal L_j$ identifies outcome-bearing fields that must not replace engineering evidence. With modality availability $a_n$ and preregistered task weights $w_j(z_n)$,
\begin{equation}
g_j(x_n)=\frac{a_n\odot w_j(z_n)}{\mathbf1^\top(a_n\odot w_j(z_n))}.
\label{eq:reference}
\end{equation}
Structural rules weight the source that defines the answer, generative rules follow the controlled signal-generation process, and conditional rules change weights only through registered metadata such as event stage. Counterfactual siblings preserve the physical state and target, $\psi(x_n)=\psi(\tilde x_n)$ and $y_n=\tilde y_n$, while relocating or contradicting selected evidence. Fractional weights permit tasks for which multiple modalities are jointly necessary; one-hot references are used only when a single source is sufficient. Reference construction is reviewed independently of model outputs, and disagreement among engineering rules is retained as a reference interval rather than collapsed post hoc. References, thresholds, and sibling rules are frozen before testing.

\begin{figure}[]
\centering
\includegraphics[width=0.6\columnwidth]{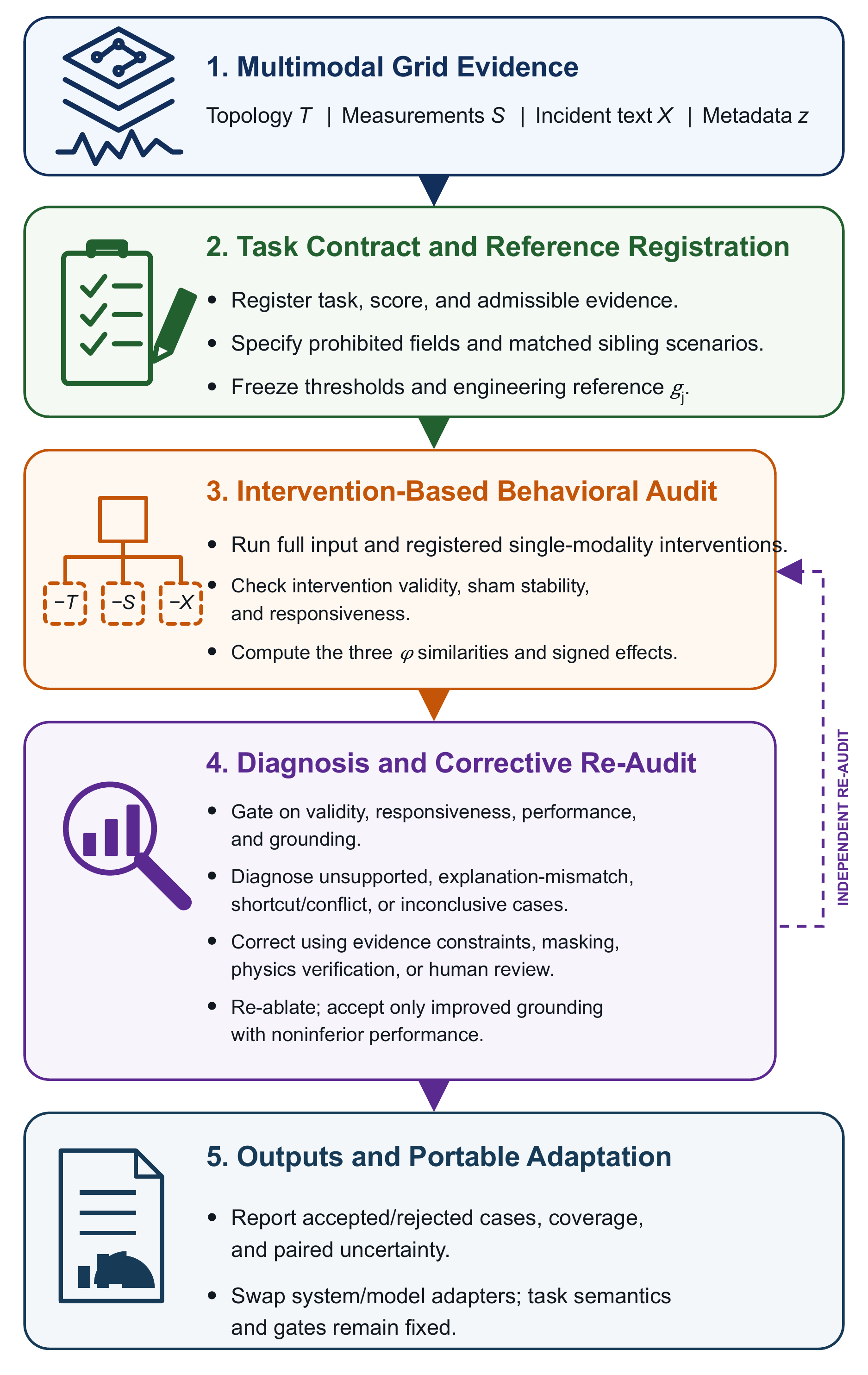}
\caption{Task-conditioned audit using full and single-modality interventions.}
\label{fig:framework}
\end{figure}

\subsection{Intervention-Based Behavioral Audit}
Matched calls evaluate
\begin{equation}
R^0_n=F_\theta(x_n),\qquad R^{-i}_n=F_\theta(A_i(x_n)),\ i=1,\ldots,K,
\label{eq:responses}
\end{equation}
where $A_i$ neutralizes only modality $i$ using an explicit availability marker and preserves the query, target, non-intervened values, schema, and decoding settings. Pair validity $I_{j,n,i}=1$ requires preserved physical invariants and valid full/ablated outputs; $I^\star_{j,n}=\prod_i I_{j,n,i}$ denotes an entirely valid intervention set. Each intervention is accompanied by a sham edit that preserves the evidence while matching formatting, which screens for prompt-length and placeholder artifacts. Positive controls confirm that the operator perturbs tasks whose registered reference assigns nonzero importance to modality $i$. Behavioral change and normalized reliance are
\begin{equation}
\delta_{j,n,i}=d_j(o_j(R^0_n),o_j(R^{-i}_n)),\quad
b_{j,n,i}=\frac{I_{j,n,i}\delta_{j,n,i}}{\sum_\ell I_{j,n,\ell}\delta_{j,n,\ell}}.
\label{eq:behavior}
\end{equation}
For categorical tasks, $d_j$ is total-variation distance when probabilities are available and a hard-label indicator otherwise; for scalar tasks, $d_j=\min(|a-a'|/\tau_j,1)$. If $r_{j,n}=\sum_i I_{j,n,i}\delta_{j,n,i}\le\tau_r$, the sample is inconclusive rather than assigned uniform reliance. Responsiveness is $\rho_j=|\{n:r_{j,n}>\tau_r\}|/|\mathcal V_j|$. Separately, the paired signed utility effect
\begin{equation}
\bar\Delta_{j,i}=N_{j,i}^{-1}\sum_n I_{j,n,i}[p_j(R^0_n)-p_j(R^{-i}_n)]
\label{eq:effect}
\end{equation}
distinguishes useful ($\bar\Delta>0$), neutral, and harmful ($\bar\Delta<0$) influence. Confidence intervals resample matched scenario families. For stochastic decoding, $L$ repeated full/ablated pairs are evaluated with identical sampling settings and $\delta$ is averaged before normalization. The resulting $b$ is interpreted as marginal dependence under the registered intervention, not as identification of the model's latent causal mechanism.

\begin{figure*}[t]
\centering
\includegraphics[width=0.75\textwidth]{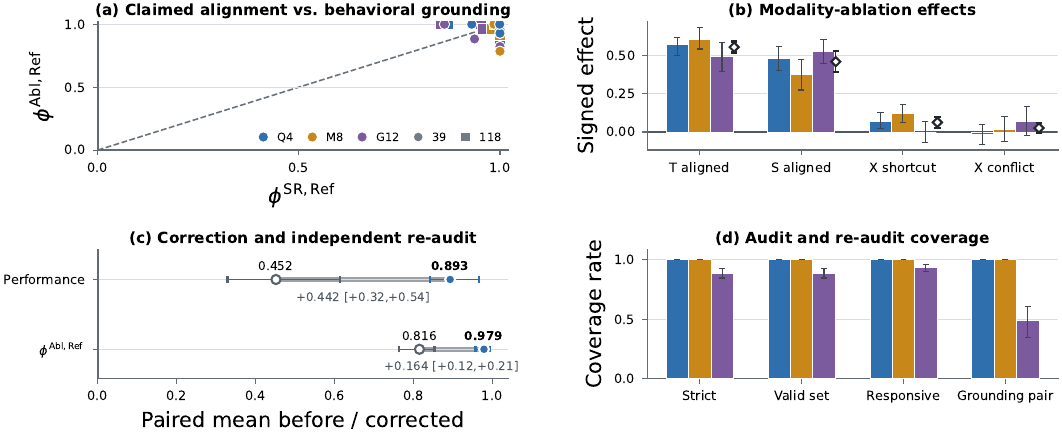}
\caption{Measured diagnostics: (a) stressed claimed alignment versus behavioral grounding; (b) signed modality effects across controlled regimes; (c) paired correction gains in performance and grounding; and (d) intervention validity, responsiveness, and audit coverage.}
\end{figure*}
\vspace{-0.5cm}
\subsection{Discrepancy Diagnosis and Corrective Re-Audit}
The measurement triangle is
\begin{align}
\phi^{\rm SR,Ref}&=\mathrm{sim}(s,g),&
\phi^{\rm Abl,Ref}&=\mathrm{sim}(b,g),\notag\\
\phi^{\rm SR,Abl}&=\mathrm{sim}(s,b),&&
\label{eq:triangle}
\end{align}
where cosine similarity measures claimed alignment, behavioral grounding, and explanation fidelity. Calibration-set thresholds distinguish grounded agreement, unsupported self-report (high SR--Ref but low Abl--Ref), explanation mismatch (high Abl--Ref but low SR--Abl), and shortcut/conflict use (low Abl--Ref interpreted with $\bar\Delta_i$). Diagnosis is assigned only when the relevant confidence interval does not cross its calibrated boundary; otherwise the case is marked ambiguous and routed for additional sampling. A response passes only when
\begin{equation}
G_{j,n}=\mathbf1\!\left\{\begin{aligned}
&V_{j,n}I^\star_{j,n}=1,\ r_{j,n}>\tau_r,\ p_{j,n}\ge\tau_P,\\[-1mm]
&\phi^{\rm Abl,Ref}_{j,n}\ge\tau_B
\end{aligned}\right\}.
\label{eq:gate}
\end{equation}
Failure triggers diagnosis-specific correction: constrain citations to $\mathcal E_j$, mask fields in $\mathcal L_j$, request evidence-linked reasoning, or invoke power-flow/contingency verification. Unsupported self-report triggers evidence citation and tool verification; explanation mismatch triggers provenance-constrained generation; shortcut/conflict use triggers masking, counterfactual prompting, or human escalation. The regenerated response is independently re-ablated and accepted only when the lower confidence bound of $\Delta\phi^{\rm Abl,Ref}$ is positive and performance is noninferior, $\Delta P=P^+-P^0_{\rm fail}\ge-\epsilon_P$.
\vspace{-0.3cm}
\subsection{Implementation and Adaptation}
The framework uses a common audit engine with two lightweight
interfaces. The system adapter maps grid data
into canonical inputs, generates registered counterfactuals, and checks
physical consistency. The model adapter handles
prompting, parsing, tool syntax, and decoding. Task definitions,
reference rules, interventions, metrics, gates, and reporting remain
fixed, so a new system or LLM requires only adapter replacement. The workflow registers tasks and thresholds, validates the adapters,
executes matched full-input and ablated calls, computes validity,
responsiveness, signed effects, and SR--Abl--Ref scores, then routes
failed cases to evidence constraints, physics tools, or human review
before independent re-auditing. Malformed outputs remain invalid, and an execution manifest records the adapters, prompts, interventions, model settings, tool outputs, and
scenario identifiers.

\section{Case Study}
We validate the framework on IEEE 39- and 118-bus systems as a proof of
concept using locally served quantized checkpoints: Qwen3 4B Instruct
(Q4), Ministral 3 8B (M8), and Gemma 3 12B (G12) \cite{ollama}.
System records are serialized into topology ($T$), measurements ($S$),
and text ($X$), covering network structure, operating conditions, and
incident narratives. The audited tasks are topology-change
identification, bottleneck localization, event classification, N--1
screening, and congestion-impact estimation. Performance $P$ is
accuracy for the first four tasks and $1-$normalized MAE for the last.
Five physical families per task--system cell yield 50 unique families,
150 model--family trials, 450 model--regime observations, and 2606
registered calls, including re-audits of 89 failed stressed cases.
Result panels report 95\% task-stratified family-bootstrap intervals
from 2,000 resamples, capturing scenario but not model-population
variation.

\subsection{Subcase A: Cross-Model Validation and Ablation}
\begin{table}[]
\caption{Cross-model tests. A/E: aligned/stressed. All $\phi$ columns are the cosine similarities in~\eqref{eq:triangle}; correction cells give paired mean change (valid-pair count)}
\label{tab:test}
\centering
\scriptsize
\setlength{\tabcolsep}{1.05pt}
\begin{tabular}{@{}llccccccc@{}}
\toprule
Mdl. & Sys. & $P_A$ & $\phi_A^{\rm Abl,Ref}$ & $P_E$ & $\phi_E^{\rm SR,Ref}$ & $\phi_E^{\rm Abl,Ref}$ & $\Delta P_c\,(n)$ & $\Delta\phi_c^{\rm Abl,Ref}\,(n)$\\
\midrule
\multirow{2}{*}{Q4} & 39 & .875 & .983 & .895 & .986 & .977 & $+.833\,(6)$ & $+.190\,(6)$\\
 & 118 & .880 & .960 & .900 & .974 & .966 & $+.500\,(10)$ & $+.172\,(10)$\\
\multirow{2}{*}{M8} & 39 & .787 & .966 & .827 & .994 & .951 & $+.555\,(15)$ & $+.163\,(15)$\\
 & 118 & .758 & .974 & .819 & .993 & .967 & $+.692\,(13)$ & $+.128\,(13)$\\
\multirow{2}{*}{G12} & 39 & .896 & .887 & .834 & .950 & .903 & $+.147\,(19)$ & $+.150\,(14)$\\
 & 118 & .875 & .963 & .844 & .951 & .949 & $+.250\,(12)$ & $+.215\,(8)$\\
\bottomrule
\end{tabular}
\end{table}

\begin{table}[]
\caption{Signed modality effects. A/Sh/Cf: aligned/shortcut/conflict.}
\label{tab:ablation}
\centering
\scriptsize
\setlength{\tabcolsep}{1.55pt}
\begin{tabular}{@{}llrrrrrr@{}}
\toprule
Mdl. & Sys. & $\Delta_T^A$ & $\Delta_S^A$ & $\Delta_X^{\rm Sh}$ & $\Delta_X^{\rm Cf}$ & $\phi_{\rm Sh}^{\rm Abl,Ref}$ & $\phi_{\rm Cf}^{\rm Abl,Ref}$\\
\midrule
\multirow{2}{*}{Q4} & 39 & $+.560$ & $+.475$ & $+.040$ & $-.080$ & .983 & .971\\
 & 118 & $+.576$ & $+.480$ & $+.093$ & $+.053$ & .965 & .966\\
\multirow{2}{*}{M8} & 39 & $+.680$ & $+.387$ & $+.160$ & $+.080$ & .931 & .971\\
 & 118 & $+.520$ & $+.358$ & $+.082$ & $-.040$ & .960 & .974\\
\multirow{2}{*}{G12} & 39 & $+.435$ & $+.548$ & $+.000$ & $+.100$ & .945 & .870\\
 & 118 & $+.542$ & $+.500$ & $+.000$ & $+.043$ & .950 & .948\\
\bottomrule
\end{tabular}
\end{table}
In this subcase, we examined cross-model agreement among performance,
claimed alignment $\phi^{\rm SR,Ref}$, behavioral grounding
$\phi^{\rm Abl,Ref}$, and task-conditioned ablation effects, and tested
correction on failed stressed cases. On jointly estimable stressed
observations,
$\phi^{\rm SR,Ref}-\phi^{\rm Abl,Ref}=0.024$ $[0.011,0.036]$.
Correction improves performance from $0.452$ to $0.893$
$(+0.442,\,[0.324,0.543];\,75/89)$ and grounding from $0.816$ to
$0.979$ $(+0.164,\,[0.124,0.213];\,66/89)$. Across models, aligned
topology and measurement effects are $+0.552$ and $+0.458$,
respectively. Shortcut text adds utility $(+0.062)$ despite having no
registered engineering importance, while conflict-text effects are near
zero overall but vary across model--system pairs. These results support
cross-model mismatch detection, correction, and signed
task-conditional diagnosis, but not uniformly harmful conflict.

\subsection{Subcase B: Gate Detection, Correction, and Audit Coverage}
\begin{table}[H]
\caption{Audit coverage and controlled-regime gate detection. $I^\star$: valid intervention set; $\rho$: responsiveness given validity; BA: balanced accuracy; Sp: specificity; $\phi_B\equiv\phi^{\rm Abl,Ref}$; pair counts use eligible failed stressed cases.}
\label{tab:audit}
\centering
\scriptsize
\setlength{\tabcolsep}{2.0pt}
\begin{tabular}{@{}lcccccc@{}}
\toprule
Mdl. & $I^\star$ & $\rho$ & BA & Sp & $n_P/n_E$ & $n_{\phi_B}/n_E$\\
\midrule
Q4 & 1.000 & 1.000 & .869 & .938 & 16/16 & 16/16\\
M8 & 1.000 & 1.000 & .943 & .885 & 28/28 & 28/28\\
G12 & .887 & .932 & .876 & .779 & 31/45 & 22/45\\
\bottomrule
\end{tabular}
\end{table}
In this subcase, we tested controlled-failure detection, intervention validity and responsiveness, and independent re-audit coverage. Correction targets shortcut or conflict observations failing~\eqref{eq:gate}; all 89 eligible failures form the denominator, and missing pairs are not imputed. Balanced accuracy is .869--.943 and specificity .779--.938. Q4/M8 attain complete validity, responsiveness, and pair coverage. G12 remains responsive when valid but retains 31/45 performance and 22/45 $\phi^{\rm Abl,Ref}$ pairs because its strict structured-output rate is .887. The audit therefore transfers cleanly to Q4/M8 while exposing G12 adapter attrition.

\section{Conclusion}
In this letter, we proposed a task-conditional faithfulness audit for multimodal LLM-based grid diagnosis. The framework jointly evaluates performance, reported and behavioral reliance, explanation fidelity, signed modality effects, and engineering reference importance. The results show that reported alignment can exceed behavioral grounding under stress, while targeted correction improves both performance and grounding. Model-specific responses to shortcut and conflicting text further motivate reporting validity and responsiveness alongside aggregate faithfulness scores. Future work will evaluate the audit in utility-scale grid workflows.

\end{document}